%% file: multibox_arxiv.tex
\documentclass[10pt,twocolumn,letterpaper]{article}

\usepackage{cvpr}
\usepackage{times}
\usepackage{epsfig}
\usepackage{graphicx}
\usepackage{amsmath}
\usepackage{amssymb}

\usepackage[pagebackref=true,breaklinks=true,letterpaper=true,colorlinks,bookmarks=false]{hyperref}

\cvprfinalcopy 

\def\cvprPaperID{XXX} 

\ifcvprfinal\fi
\begin{document}

\title{Scalable Object Detection using Deep Neural Networks}

\author{Dumitru Erhan\ \ \ Christian Szegedy\ \ \ Alexander Toshev\ \ \ Dragomir Anguelov\\
Google \\
{\tt\small {\{dumitru, szegedy, toshev, dragomir\}}@google.com}
}

\maketitle

\begin{abstract} Deep convolutional neural networks have recently achieved
  state-of-the-art performance on a number of image recognition benchmarks,
  including the ImageNet Large-Scale Visual Recognition Challenge
  (ILSVRC-2012).  The winning model on the localization sub-task was a network
  that predicts a single bounding box and a confidence score for each object
  category in the image. Such a model captures the whole-image context around
  the objects but cannot handle multiple instances of the same object in the
  image without naively replicating the number of outputs for each instance.
  In this work, we propose a saliency-inspired neural network model for
  detection, which predicts a set of class-agnostic bounding boxes along with a
  single score for each box, corresponding to its likelihood of containing
  \textit{any} object of interest. The model naturally handles a variable
  number of instances for each class and allows for cross-class generalization
  at the highest levels of the network. We are able to obtain competitive
  recognition performance on VOC2007 and ILSVRC2012, while using only the top
  few predicted locations in each image and a small number of neural network
  evaluations.  \end{abstract}

\input{intro}
\input{previous_work}
\input{proposed_approach}
\input{trainingeval}
\input{voc_results}
\input{imagenet_results}

\input{conclusion}

{\small
\bibliographystyle{ieee}
\bibliography{multibox_arxiv}
}

\end{document}

%% file: intro.tex
\section{Introduction}
Object detection is one of the fundamental tasks in computer vision. A common
paradigm to address this problem is to train object detectors which operate on
a subimage and apply these detectors in an exhaustive manner across all locations
and scales. This paradigm was successfully used within a discriminatively trained
Deformable Part Model (DPM) to achieve state-of-art results on detection tasks
\cite{felzenszwalb2010object}.

The exhaustive search through all possible locations and scales poses a computational
challenge. This challenge becomes even harder as the number of classes grows, since
most of the approaches train a separate detector per class. In order to address this issue
a variety of methods were proposed, varying from detector cascades, to using segmentation
to suggest a small number of object hypotheses
\cite{van2011segmentation,carreira2010constrained,endres2010category}.

In this paper, we ascribe to the latter philosphy and propose to train a detector, called ``DeepMultiBox'','
which generates a few bounding boxes as object candidates. These boxes are
generated by a \textit{single} DNN in a \textit{class agnostic} manner. Our model
has several contributions. First, we define object detection as a regression problem
to the coordinates of several bounding boxes. In addition, for each predicted box
the net outputs a confidence score of how likely this box contains an object.
This is quite different from traditional approaches, which
score features within predefined boxes, and has the advantage of expressing
detection of objects in a very compact and efficient way.

The second major contribution is the loss, which trains the bounding box predictors
as part of the network training. For each training example, we solve an assignment
problem between the current predictions and the groundtruth boxes and update the matched
box coordinates, their confidences and the underlying features through Backpropagation. In this
way, we learn a deep net tailored towards our localization problem. We
capitalize on the excellent representation learning abilities of DNNs, as recently
exeplified recently in image classification \cite{krizhevsky2012imagenet} and
object detection settings \cite{szegedy2013detection}, and perform joint learning of
representation and predictors.

Finally, we train our object box predictor in a class-agnostic manner. We consider this
as a scalable way to enable efficient detection of large number of object classes. We
show in our experiments that by only post-classifying less than ten boxes, obtained by
a single network application, we can achieve state-of-art detection results. Further, we
show that our box predictor generalizes over unseen classes and as such is flexible to be
re-used within other detection problems.

%% file: previous_work.tex
\section{Previous work}
The literature on object detection is vast, and in this section we will
focus on approaches exploiting class-agnostic ideas and addressing scalability.

Many of the proposed detection approaches are based on part-based models \cite{fischler1973representation},
which more recently have achieved
impressive performance thanks to discriminative learning and carefully
crafted features \cite{felzenszwalb2010object}. These methods, however, rely on
exhaustive application of part templates over multiple scales and as such are expensive.
Moreover, they scale linearly in the number of classes, which becomes a
challenge for modern datasets such as ImageNet.

To address the former issue, Lampert et al.~\cite{lampert2008beyond} use a
branch-and-bound strategy to avoid evaluating all potential object locations.
To address the latter issue, Song et al.~\cite{song2012sparselet} use a
low-dimensional part basis, shared across all object classes. A hashing based
approach for efficient part detection has shown good results as well
\cite{dean2013fast}.

A different line of work, closer to ours, is based on the idea that objects
can be localized \textit{without} having to know their class. Some
of these approaches build on bottom-up classless segmentation \cite{gu2009recognition}.
The segments, obtained in this way, can be scored using top-down feedback
\cite{van2011segmentation,carreira2010constrained,endres2010category}.  Using the same
motivation, Alexe et al.~\cite{alexe2010object} use an inexpensive classifier to
score object hypotheses for being an object or not and in this way reduce the
number of location for the subsequent detection steps. These
approaches can be thought of as Multi-layered models, with segmentation as first
layer and a segment classification as a subsequent layer. Despite the fact that
they encode proven perceptual principles, we will show that having deeper
models which are fully learned can lead to superior results.

Finally, we capitalize on the recent advances in Deep Learning, most noticeably
the work by Krizhevsky et al.~\cite{krizhevsky2012imagenet}. We extend their
bounding box regression approach for detection to the case of handling
multiple objects in a scalable manner. DNN-based regression, to object
masks however, has been applied by Szegedy et al.~\cite{szegedy2013detection}. 
This last approach achieves state-of-art detection performance but does not 
scale up to multiple classes due to the cost of a single mask regression.





%% file: proposed_approach.tex
\section{Proposed approach}
We aim at achieving a class-agnostic scalable object detection by predicting
a set of bounding boxes, which represent potential objects. More precisely, we
use a Deep Neural Network (DNN), which outputs a fixed number of bounding boxes.
In addition, it outputs a score for each box expressing the networkconfidence of
this box containing an object.

\vspace{-0.5cm}
\paragraph{Model} To formalize the above idea, we encode the $i$-th object box and its
associated confidence as node values of the last net layer:
\begin{description}
\item[Bounding box:] we encode the upper-left and lower-right coordinates of
each box as four node values, which can be written as a vector
$l_i\in\mathbb{R}^4$. These coordinates are normalized w.~r.~t.~image dimensions
to achieve invariance to absolute image size. Each normalized coordinate is
produced by a linear transformation of the last hidden layer.
\item[Confidence:] the confidence score for the box containing an object is encoded as
a single node value $c_i\in[0, 1]$. This value is produced through a linear transformation
of the last hidden layer followed by a sigmoid.
\end{description}
We can combine the bounding box locations $l_i$, $i\in\{1,\ldots K\}$, as one
linear layer. Similarly, we can treat collection of all confidences $c_i$,
$i\in\{1,\ldots K\}$ as the output as one sigmoid layer. Both these output layers are
connected to the last hidden layers.

At inference time, out algorithm produces $K$ bounding boxes. In our experiments,
we use $K=100$ and $K=200$. If desired, we can use the confidence scores and non-maximum
suppression to obtain a smaller
number of high-confidence boxes at inference time. These boxes
are supposed to represent objects. As such, they can be classified with a
subsequent classifier to achieve object detection. Since the number of boxes is
very small, we can afford powerful classifiers. In our
experiments, we use another DNN for classification \cite{krizhevsky2012imagenet}.

\vspace{-0.5cm}
\paragraph{Training Objective} We train a DNN to predict bounding boxes and
their confidence scores for each training image such that the highest scoring
boxes match well the ground truth object boxes for the image.
Suppose that for a particular training example, $M$
objects were labeled by bounding boxes $g_j$, $j\in\{1,\ldots, M\}$. In
practice, the number of predictions $K$ is much larger than the number of
groundtruth boxes $M$. Therefore, we try to optimize only the subset of predicted
boxes which match best the ground truth ones. We optimize their locations to
improve their match and maximize their confidences. At the
same time we minimize the confidences of the remaining predictions, which are
deemed not to localize the true objects well.

To achieve the above, we formulate an assignment problem
for each training example. We $x_{ij}\in\{0,1\}$ denote the assignment:
$x_{ij} = 1$ iff the $i$-th prediction is assigned to $j$-th true object. The
objective of this assignment can be expressed as:
\begin{equation}\label{eq:assignment_objective}
  F_{\text{match}}(x, l)=\frac{1}{2}\sum_{i,j}x_{ij}||l_i-g_j||_2^2
\end{equation}
where we use $L_2$ distance between the normalized bounding box coordinates to
quantify the dissimilarity between bounding boxes.

Additionally, we want to optimize the confidences of the boxes according to the
assignment $x$. Maximizing the confidences of assigned predictions
can be expressed as:
\begin{equation}\label{eq:confidence_objective}
F_{\text{conf}}(x, c)=-\sum_{i,j}x_{ij}\log(c_i) - \sum_i(1-\sum_jx_{ij})\log(1-c_i)
\end{equation}
In the above objective $\sum_jx_{ij} =1$ iff prediction $i$ has been matched to
a groundtruth. In that case $c_i$ is being maximized, while in the opposite case
it is being minimized. A different interpretation of the above term is achieved
if we $\sum_jx_{ij}$ view as a probability of prediction $i$ containing an object of interest. Then, the
above loss is the negative of the entropy and thus corresponds to a max entropy loss.

The final loss objective combines the matching and confidence losses:
\begin{equation}
F(x,l,c) = \alpha F_{\text{match}}(x, l) + F_{\text{conf}}(x, c)
\end{equation}
subject to constraints in Eq.~\ref{eq:assignment_objective}. $\alpha$ balances the contribution of the different loss terms.
\vspace{-0.5cm}
\paragraph{Optimization} For each training example, we solve for an optimal assignment $x^*$ of predictions to
true boxes by
\begin{eqnarray}
x^*& =&\arg\min_xF(x,l,c) \\
\text{subject to } && x_{ij}\in\{0,1\}, \sum_ix_{ij} = 1,
\end{eqnarray}
where the constraints enforce an assignment solution. This is a variant of bipartite matching,
which is polynomial in complexity. In our application the matching is very inexpensive -- the number of labeled
objects per image is less than a dozen and in most cases only very few objects are labeled.

Then, we optimize the network parameters via back-propagation. For example,
the first derivatives of the back-propagation
algorithm are computed w.~r.~t.~ $l$ and $c$:
\begin{eqnarray}
  \frac{\partial F}{\partial l_i} & =&\sum_j(l_i - g_j)x_{ij}^* \\
  \frac{\partial F}{\partial c_i} & = & \frac{\sum_jx_{ij}^*c_i}{c_i(1-c_i)}
\end{eqnarray}

\vspace{-0.5cm}\paragraph{Training Details} While the loss as defined above is
in principle sufficient, three
modifications make it possible to reach better accuracy significantly faster.
The first such modification is to perform clustering of ground truth locations
and find $K$ such clusters/centroids that we can use as priors for each of the
predicted locations. Thus, the learning algorithm is encouraged to learn a
residual to a prior, for each of the predicted locations.

A second modification pertains to using these priors in the matching
process: instead of
matching the $N$ ground truth locations with the $K$ predictions, we find the best
match between the $K$ priors and the ground truth. Once the matching is done, the
target confidences are computed as before. Moreover, the location prediction
loss is also unchanged: for any matched pair of (target, prediction) locations,
the loss is defined by the difference between the groundtruth and the
coordinates that correspond to the matched prior. We call the usage of priors
for matching \emph{prior matching} and hypothesize that it enforces diversification among
the predictions.

It should be noted, that although we defined our method in a class-agnostic
way, we can apply it to predicting object boxes for a particular class. To do this,
we simply need to train our models on bounding boxes for that class.

Further, we can predict $K$ boxes per
class. Unfortunately, this model will have number of parameters growing
linearly with the number of classes. Also, in a typical setting, where the number of
objects for a given class is relatively small, most of these parameters will
see very few training examples with a corresponding gradient contribution. We
argue thusly that our two-step process -- first localize, then recognize -- is
a superior alternative in that it allows leveraging data from multiple object
types in the same image using a small number of parameters.

\section{Experimental results}

%% file: trainingeval.tex
\subsection{Network Architecture and Experiment Details}

The network architecture for the localization and classification
models that we use is the same as the one used by~\cite{krizhevsky2012imagenet}.
We use Adagrad for controlling the learning rate decay, mini-batches of
size 128, and parallel distributed training with multiple identical replicas
of the network, which achieves faster convergence. As mentioned previously,
we use priors in the localization loss -- these are computed using $k$-means
on the training set. We also use an $\alpha$
of $0.3$ to balance the localization and confidence losses.

The localizer might output coordinates outside the crop area used for the
inference. The coordinates are mapped and truncated to the final image area,
at the end. Boxes are additionally pruned using non-maximum-suppression with
a Jaccard similarity threshold of $0.5$. Our second model then classifies
each bounding box as objects of interest or ``background''.

To train our localizer networks, we generated approximately $30$ million images from the
training set, applying the following procedure to each image in the training set.
The samples are shuffled at the end.
To train our localizer networks, we generated approximately $30$ million images
from the training set by applying the following procedure to each image in the
training set.
For each image, we generate the same number of square samples such that the 
total number of samples is about ten million.
For each image, the samples are bucketed such that for each of the ratios in
the ranges of $0-5\%, 5-15\%, 15-50\%, 50-100\%$,
there is an equal number of samples in which the ratio covered by the bounding
boxes is in the given range.

The selection of the training set and most of our hyper-parameters were based on past
experiences with non-public data sets. For the experiments below we have not
explored any non-standard data generation or regularization options.

In all experiments, all hyper-parameters were selected by evaluating on a
held out portion of the training set (10\% random choice of examples).

%% file: voc_results.tex
\subsection{VOC 2007}

The Pascal Visual Object Classes (VOC) Challenge \cite{everingham2010pascal}
is the most commong benchmark for object detection algorithms. It consists mainly
of complex scene images in which bounding boxes of 20 diverse object classes
were labelled.

In our evaluation we focus on the 2007 edition of VOC, for which a test set was
released. We present results by training on VOC 2012, which contains approx.~11000 images.
We trained a $100$ box localizer as well as a deep net based classifier
\cite{krizhevsky2012imagenet}.

\subsubsection{Training methodology}
We trained the classifier on a data set comprising of
\begin{itemize}
\item $10$ million crops overlapping some object with at least $0.5$ Jaccard overlap
similarity. The crops are labeled with one of the 20 VOC object classes.
\item $20$ million negative crops that have at most $0.2$ Jaccard similarity
with any of the object boxes. These crops are labeled with the special
``background'' class label.
\end{itemize}
The architecture and the selection of hyperparameters followed that of
\cite{krizhevsky2012imagenet}.

\subsubsection{Evaluation methodology}

In the first round, the localizer model is applied to the maximum center
square crop in the image. The crop is resized to the network input size which is
$220 \times 220$. A single pass through this network gives us up to hundred
candidate boxes. After a non-maximum-suppression with overlap threshold $0.5$,
the top $10$ highest scoring detections are kept and were classified by the
21-way classifier model in a separate passes through the network.
The final detection score is the product of the localizer score for the given
box multiplied by the score of the classifier evaluated on the maximum square
region around the crop. These scores are passed to the evaluation and were used
for computing the precision recall curves.

\subsection{Discussion}

\begin{figure}
{\centering
\includegraphics[width=0.4\textwidth]{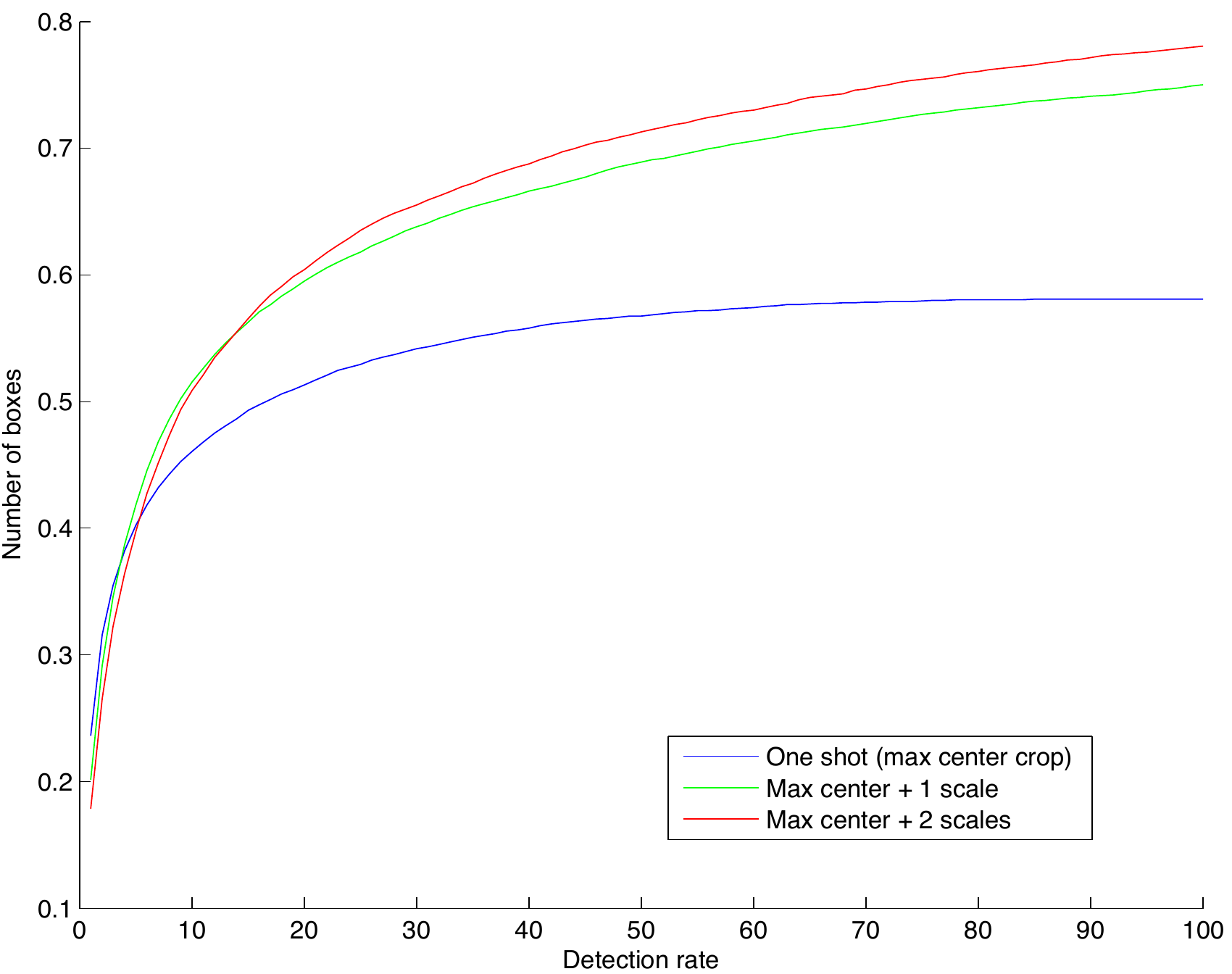}
\caption{\label{fig:saliency}Detection rate of class ``object''
vs number of bounding boxes per image. The model, used for these results, was trained on
VOC 2012.}
}
\end{figure}

First, we analyze the performance of our localizer in isolation. We present the number of detected objects,
as defined by the Pascal detection criterion,
against the number of produced bounding boxes. In Fig.~\ref{fig:saliency} plot we show results obtained by
training on VOC2012. In addition, we present
results by using the max-center square crop of the image as input as well as by
using two scales: the max-center crop by a second scale where we select $3\times 3$ windows
of size $60\%$ of the image size.

As we can see, when using a budget of $10$
bounding boxes we can localize $45.3\%$ of the objects with the first model, and $48\%$ with
the second model. This shows better perfomance than other reported results, such as
the objectness algorithm achieving $42\%$ \cite{alexe2010object}. Further, this plot
shows the importance of looking at the image at several resolutions. Although our algorithm
manages to get large number of objects by using the max-center crop, we obtain an additional boost
when using higher resolution image crops.

\begin{table*}[t]
{\centering
\begin{tabular}{|l|c|c|c|c|c|c|c|c|c|c|}
\hline
class & aero & bicycle & bird & boat & bottle & bus & car & cat & chair & cow   \\
\hline\hline
DeepMultiBox & {\bf .413} & .277 & {\bf .305} & {\bf .176} & .032 & .454 & .362 & {\bf .535} & .069 & .256 \\
\hline
3-layer model \cite{zhu2010latent} & .294 & .558 & .094 & .143 & .286 &  .440 & .513 & .213 & .200 & .193  \\
\hline
Felz. et al. \cite{felzenszwalb2010object} & .328 & .568 & .025 & .168 & {\bf .285} & .397 & .516 & .213 & .179 & .185  \\
\hline
Girshick et al.~\cite{voc-release5}  & .324 & {\bf .577} & .107 & .157 & .253 & .513 & {\bf .542} & .179 & {\bf .210} & .240 \\
\hline
Szegedy et al.~\cite{szegedy2013detection} & .292 &  .352 &  .194 & .167 & .037 & {\bf .532} & .502 & .272 & .102 &  {\bf .348}\\
\hline
\hline
class & table & dog & horse & m-bike & person & plant & sheep & sofa & train & tv \\
\hline\hline
DeepMultiBox & .273 & {\bf .464} & .312 & .297 & .375 & .074 & .298 & .211 & .436 & .225 \\
\hline
3-layer model \cite{zhu2010latent}  & .252 & .125 & .504&  .384 & .366 & {\bf .151} & .197&  .251&  .368 &  .393 \\
\hline
Felz. et al. \cite{felzenszwalb2010object} & .259&  .088 & .492 & .412 & .368 & .146 & .162 & .244 & .392 & .391 \\
\hline
Girshick et al.~\cite{voc-release5} & .257&  .116 & {\bf .556} & {\bf .475} & {\bf .435} & .145 & .226 & {\bf .342} & {\bf .442} & .413 \\
\hline
Szegedy et al~.\cite{szegedy2013detection} & {\bf .302} &  .282 & .466 & .417 & .262 & .103 &  {\bf .328} & .268 &  .398 &  {\bf .47} \\
\hline
\end{tabular}
\caption{\label{table:pr_voc} Average Precision on VOC 2007 test of our method, called DeepMultiBox, and
other competitive methods. DeepMultibox was trained on VOC2012 training data, while the rest of the models
were trained on VOC2007 data.}
}
\end{table*}

Further, we classify the produced bounding boxes by a 21-way classifier, as described above.
The average precisions (APs) on VOC 2007 are presented in Table~\ref{table:pr_voc}. The achieved
mean AP is $0.29$, which is on par with state-of-art. Note that, our running time complexity
is very low -- we simply use the top $10$ boxes.

\begin{figure*}
\includegraphics[width=1.0\textwidth]{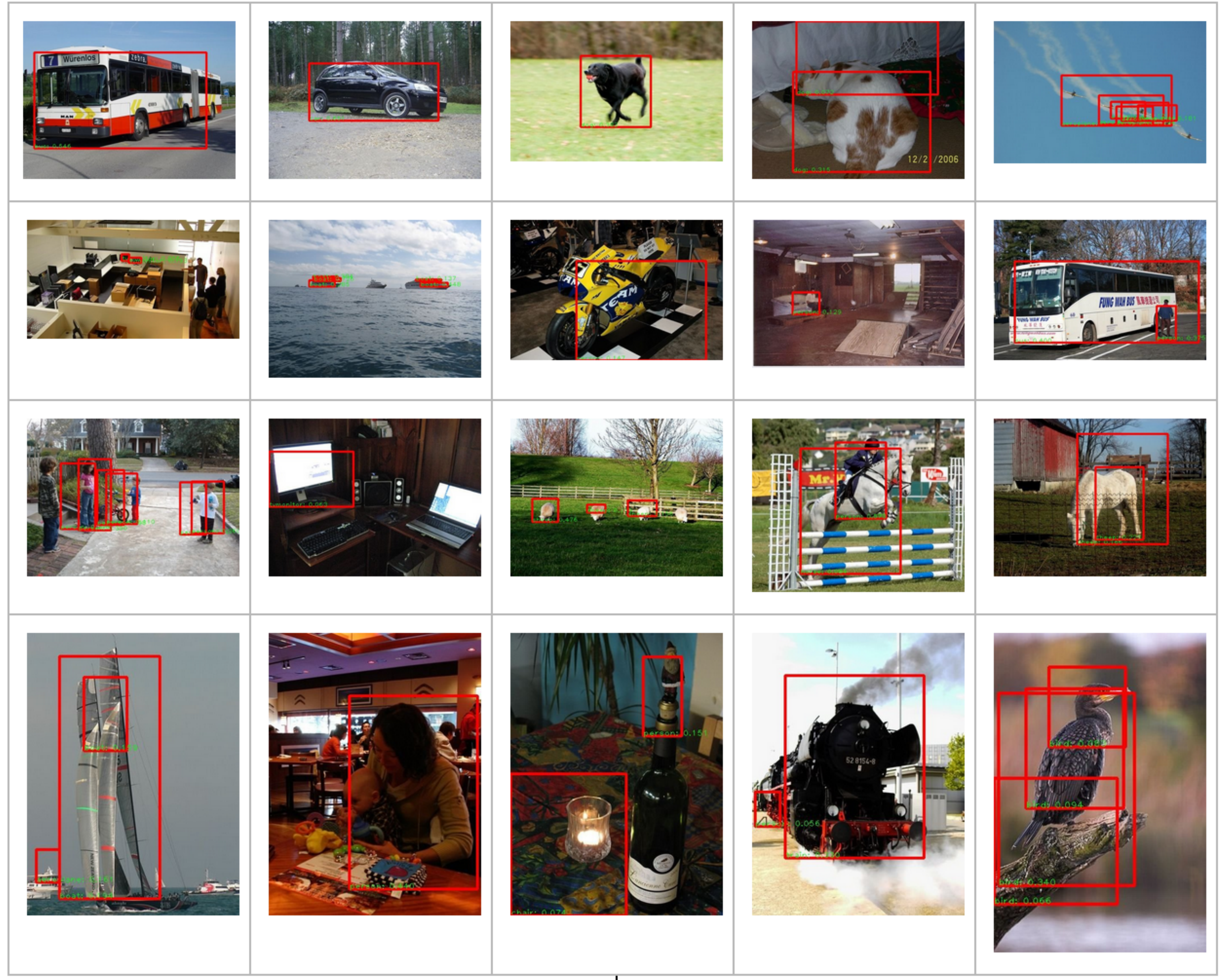}
\caption{\label{fig:vocdetections} Sample of detection results on VOC 2007.}
\end{figure*}

\begin{figure*}
\includegraphics[width=1.0\textwidth]{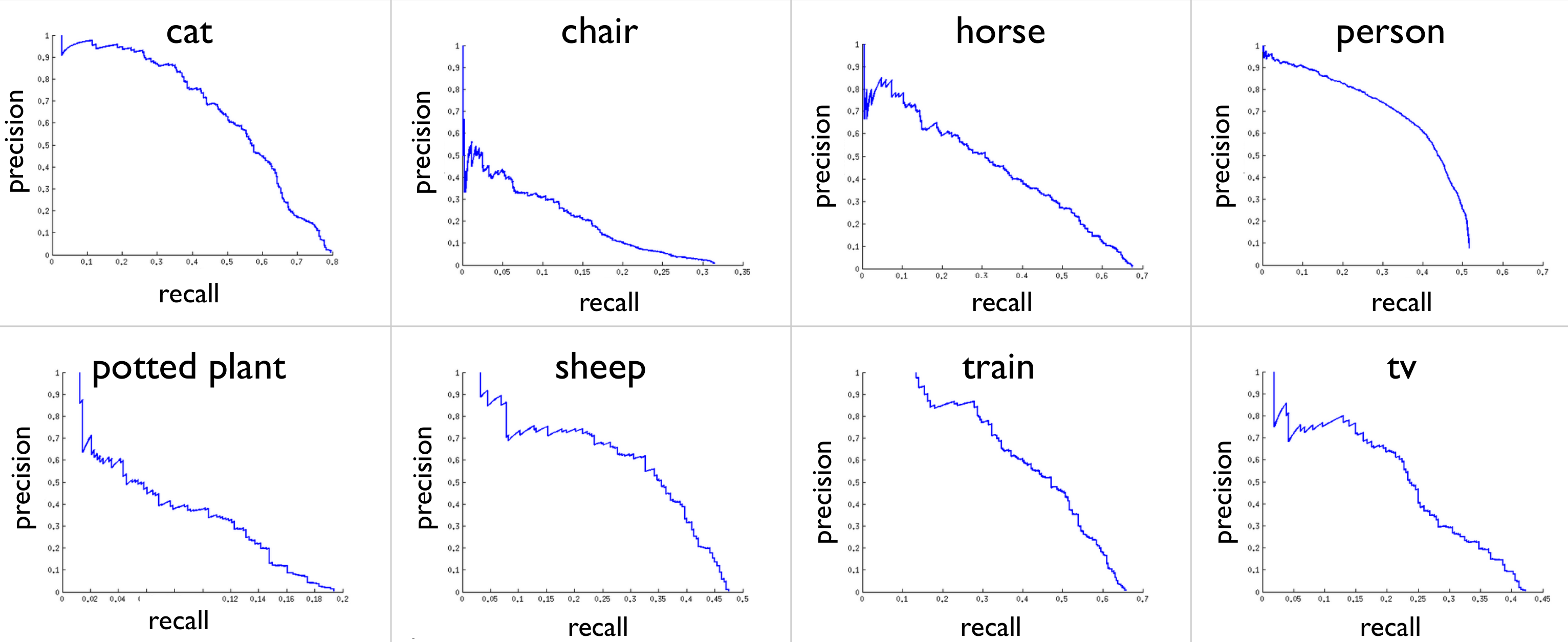}
\caption{\label{fig:vocprecrecall} Precision-recall curves on selected VOC classes.}
\end{figure*}

Example detections and full precision recall curves are shown in Fig.~\ref{fig:vocdetections}
and Fig.~\ref{fig:vocprecrecall} respectively. It is important to note that the visualized detections
were obtained by using only the max-centered square image crop, i.~e.~the full image was used.
Nevertheless, we manage to obtain relatively small objects, such as the boats in row 2 and column 2,
as well as the sheep in row 3 and column 3.

%% file: imagenet_results.tex
\subsection{ILSVRC 2012 Detection Challenge}

For this set of experiments, we used the ILSVRC 2012 detection challenge
dataset. This dataset consists of 544,545 training images labeled with
categories and locations of 1,000 object categories, relatively uniformly
distributed among the classes. The validation set, on which the performance
metrics are calculated, consists of 48,238 images. 

\subsubsection{Training methodology}

In addition to a localization model that is
identical (up to the dataset on which it is trained on) to the VOC model, we
also train a model on the ImageNet Classification challenge data, which will
serve as the recognition model. This model is trained in a procedure that is
substantially similar to that of \cite{krizhevsky2012imagenet} and is able to
achieve the same results on the classification challenge validation set; note
that we only train a single model, instead of $7$ -- the latter brings
substantial benefits in terms of classification accuracy, but is $7\times$ more
expensive, which is not a negligible factor.

Inference is done as with the VOC setup: the number of predicted locations is
$K=100$, which are then reduced by Non-Max-Suppression (Jaccard overlap
criterion of $0.4$) and which are post-scored by the classifier: the score is
the product of the localizer confidence for the given box multiplied by the
score of the classifier evaluated on the minimum square region around the crop.
The final scores (detection score times classification score) are then sorted
in descending order and only the top scoring score/location pair is kept for a
given class (as per the challenge evaluation criterion).

In all experiments, the hyper-parameters were selected by evaluating on a held
out portion of the training set (10\% random choice of examples).

\subsubsection{Evaluation methodology}
The official metric of the ``Classification with localization`` ILSVRC-2012
challenge is detection@5, where an algorithm is only allowed to produce one box
per each of the 5 labels (in other words, a model is neither penalized nor
rewarded for producing valid multiple detections of the same class), where the
detection criterion is 0.5 Jaccard overlap with any of the ground-truth boxes
(in addition to the matching class label). 

Table~\ref{t:imagenet_results} contains a comparison of the proposed method,
dubbed DeepMultiBox, with classifying the ground-truth boxes directly and with
the approach of inferring one box per class directly. The metrics reported are
detection\@5 and classification\@5, the official metrics for the ILSVRC-2012
challenge metrics. In the table, we vary the number of windows at which we
apply the classifier (this number represents the top windows chosen after
non-max-suppression, the ranking coming from the confidence scores). The
one-box-per-class approach is a careful re-implementation of the winning entry
of ILSVRC-2012 (the ``classification with localization'' challenge), with 1
network trained (instead of 7).

\begin{table}[h]
  \label{t:imagenet_results}
  \caption{Performance of Multibox (the proposed method) vs.~classifying
  ground-truth boxes directly and predicting one box per class}
  \begin{tabular}{|l|c|c|} \hline
  \bf{Method} & \bf{det@5} & \bf{class@5} \\ 
  \hline  One-box-per-class & 61.00\%    & 79.40\% \\ \hline
 Classify GT directly & 82.81\% & 82.81\% \\ \hline\hline
 DeepMultiBox, top 1 window & 56.65\% & 73.03\% \\ \hline
 DeepMultiBox, top 3 windows  & 58.71\% & 77.56\% \\ \hline 
 DeepMultiBox, top 5 windows &   58.94\% & 78.41\% \\ \hline 
 DeepMultiBox, top 10 windows &  59.06\% & 78.70\%  \\ \hline 
 DeepMultiBox, top 25 windows &   59.04\% & 78.76\% \\ \hline
  \end{tabular}
\end{table}

\begin{figure*}
\centering
\includegraphics[width=0.6\textwidth]{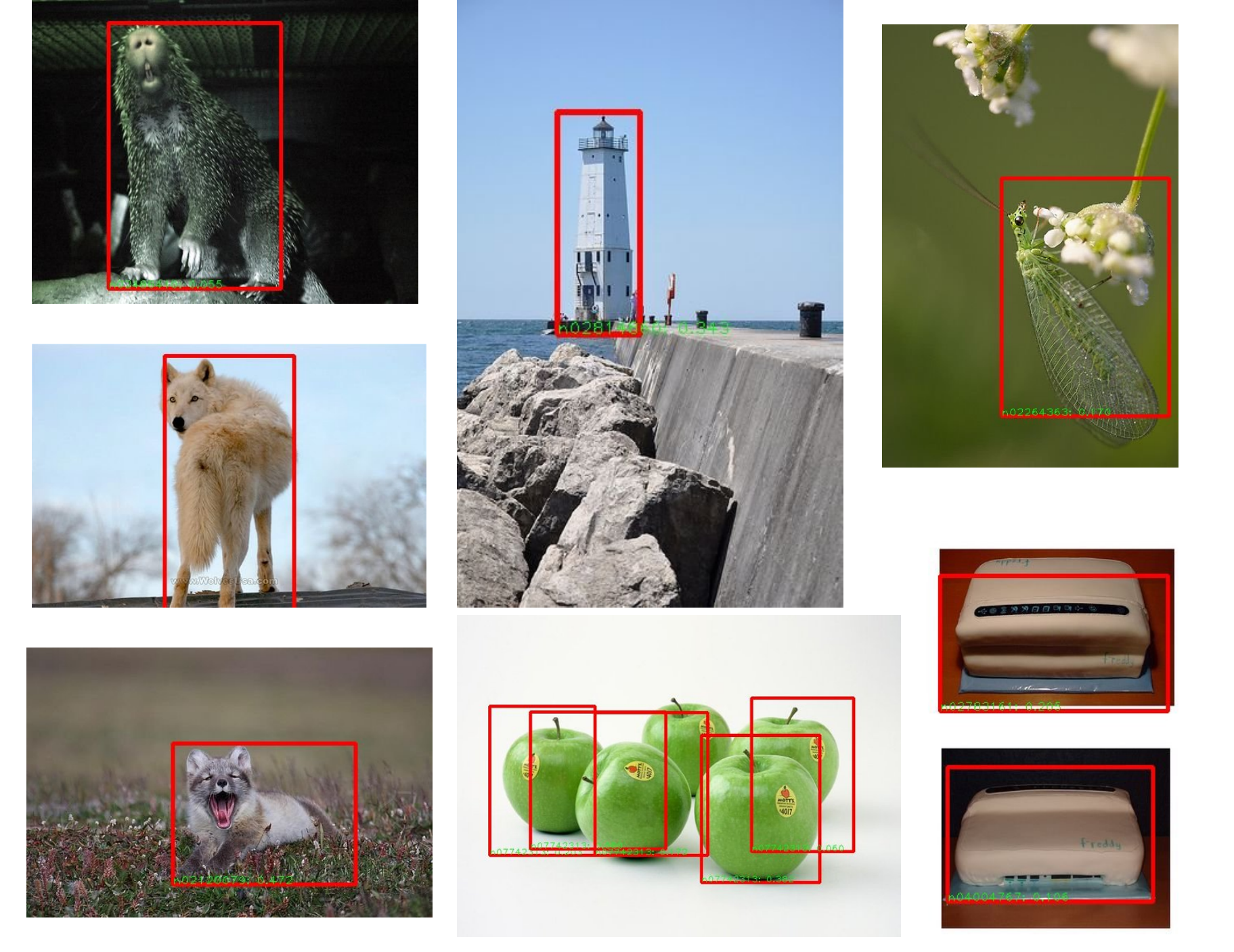}
\caption{\label{fig:imagenet} Some selected detection results on the ILSVRC-2012 detection challenge validation set.}
\end{figure*}

\begin{figure*}
\centering
\includegraphics[width=0.48\textwidth]{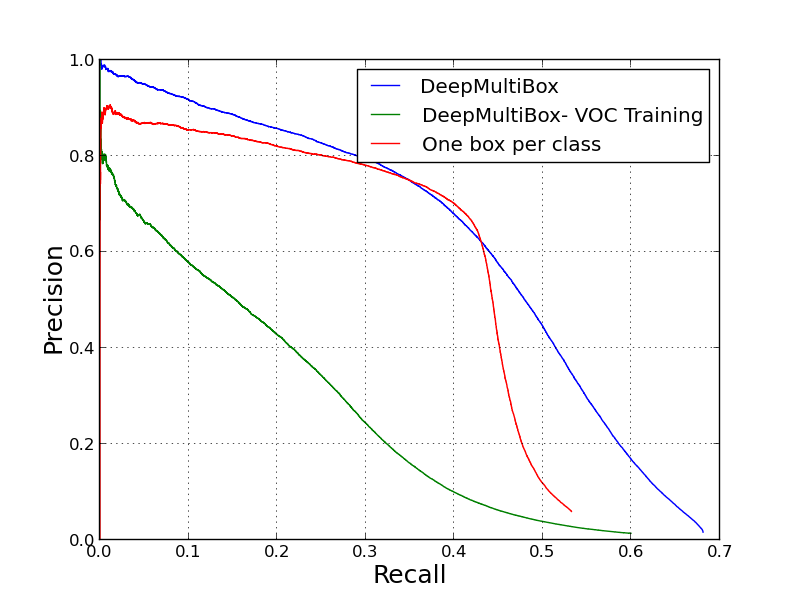}
\includegraphics[width=0.48\textwidth]{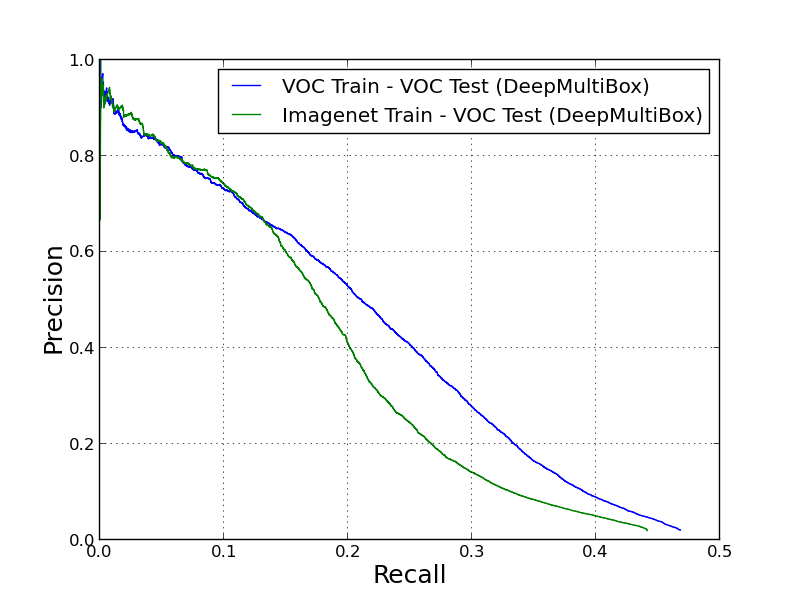}
\caption{\label{fig:agnostic_transfer} Class-agnostic detection on ILSVRC-2012 (left) and VOC 2007 (right).}
\end{figure*}

We can see that the DeepMultiBox approach is quite competitive: with 5-10
windows, it is able to perform about as well as the competing approach.  While
the one-box-per-class approach may come off as more appealing in this
particular case in terms of the raw performance, it suffers from a number of
drawbacks: first, its output scales linearly with the number of classes, for
which there needs to be training data. The multibox approach can in principle
use transfer learning to detect certain types of objects on which it has never
been specifically trained on, but which share similarities with objects that it
\emph{has} seen\footnote{For instance, if one trains with fine-grained
  categories of dogs, it will likely generalize to other kinds of breeds by
itself}.  Figure \ref{fig:agnostic_transfer} explores this hypothesis by
observing what happens when one takes a localization model trained on ImageNet
and applies it on the VOC test set, and vice-versa. The figure shows a
precision-recall curve: in this case, we perform a \emph{class-agnostic}
detection: a true positive occurs if two windows (prediction and groundtruth)
overlap by more than 0.5, independently of their class. Interestingly, the
ImageNet-trained model is able to capture more VOC windows than vice-versa: we
hypothesize that this is due to the ImageNet class set being much richer than
the VOC class set.

Secondly, the one-box-per-class approach does not generalize naturally to
multiple instances of objects of the same type (except via the the method
presented in this work, for instance). Figure \ref{fig:agnostic_transfer} shows
this too, in the comparison between DeepMultiBox and the one-per-class
approach\footnote{In the case of the one-box-per-class method,
non-max-suppression is performed on the 1000 boxes using the same criterion as
the DeepMultiBox method}. Generalizing to such a scenario is necessary for
actual image understanding by algorithms, thus such limitations need to be
overcome, and our method is a scalable way of doing so.  Evidence supporting
this statement is shown in Figure \ref{fig:agnostic_transfer} shows that the
proposed method is able to generally capture more objects more accurately that
a single-box method.

%% file: conclusion.tex
\section{Discussion and Conclusion}

In this work, we propose a novel method for localizing objects in an image,
which predicts multiple bounding boxes at a time. The method uses a deep
convolutional neural network as a base feature extraction and learning model.
It formulates a multiple box localization cost that is able to take advantage
of variable number of groundtruth locations of interest in a given image and
learn to predict such locations in unseen images.

We present results on two challenging benchmarks, VOC2007 and ILSVRC-2012, on which the proposed method
is competitive. Moreover, the method is able to perform well by
predicting only very few locations to be probed by a subsequent classifier. Our
results show that the DeepMultiBox approach is scalable and can even generalize
across the two datasets, in terms of being able to predict locations of
interest, even for categories on which it was not trained on. Additionally, it is
able to capture multiple instances of objects of the same class, which is an important
feature of algorithms that aim for better image understanding.

In the future, we hope to be able to fold the localization and recognition
paths into a single network, such that we would be able to extract both
location and class label information in a one-shot feed-forward pass through
the network. Even in its current state, the two-pass procedure (localization
network followed by categorization network) entails 5-10 network evaluations,
each at roughly 1 CPU-sec (modern machine). Importantly, this number does
\emph{not} scale linearly with the number of classes to be recognized, which
makes the proposed approach very competitive with DPM-like approaches.